\documentclass{article}

\usepackage[preprint]{corl_2025} %

\usepackage{graphicx}
\usepackage[T1]{fontenc}
\usepackage{epsfig}
\usepackage{amsmath}
\usepackage{amssymb}
\usepackage{color, soul, colortbl}
\usepackage{mathtools}
\usepackage{booktabs}
\usepackage{times}
\usepackage{microtype}
\usepackage[table,xcdraw,dvipsnames]{xcolor}
\usepackage[labelfont=bf, font=footnotesize]{caption}
\usepackage[labelfont=bf, font=footnotesize]{subcaption}
\usepackage[utf8]{inputenc}
\usepackage{float}
\usepackage{inconsolata}
\usepackage{multirow}
\usepackage{placeins}
\usepackage{stfloats}
\usepackage{enumitem}
\usepackage{tabularx}
\usepackage{xstring}
\usepackage{multirow}
\usepackage{xspace}
\usepackage{url}
\usepackage{hyperref}
\usepackage{duckuments}
\usepackage{svg}
\usepackage[hang,flushmargin]{footmisc}
\usepackage{url}
\usepackage{tabularray}
\usepackage{wrapfig}
\usepackage{epigraph}
\usepackage{amssymb}

\makeatletter
\patchcmd{\epigraph}{\@epitext{#1}}{\footnotesize\itshape\@epitext{#1}}{}{}
\makeatother

\newcommand{\model}{\texttt{Sparsh-skin}\xspace}

\newcommand{\fig}[1]{Figure~\ref{#1}}

\newcommand{\R}[1]{{%
    \textbf{%
        \ifstrequal{#1}{1}{\textcolor{red}{R#1}}{%
        \ifstrequal{#1}{2}{\textcolor{blue}{R#1}}{%
        \ifstrequal{#1}{3}{\textcolor{magenta}{R#1}}{%
        \ifstrequal{#1}{4}{\textcolor{teal}{R#1}}{%
                           \textcolor{cyan}{R#1}%
        }}}}%
    }%
}}

\definecolor{tabfirst}{rgb}{0.7, 1.0, 0.7} %
\definecolor{tabsecond}{rgb}{1, 1, 0.7} %
\definecolor{tabthird}{rgb}{1, 0.85, 0.7} %

\newcommand{\raisemath}[1]{\mathpalette{\raisemath{#1}}}

\newcommand{\av}{\mathbf{a}}

\newcommand{\pv}{\mathbf{p}}

\newcommand{\xv}{\mathbf{x}}

\newcommand{\Ev}{\mathbf{E}}

\newcommand{\Iv}{\mathbf{I}}

\newcommand{\Pv}{\mathbf{P}}

\newcommand{\Tv}{\mathbf{T}}

\newcommand{\norm}[1]{\left\lVert#1\right\rVert}

\newcommand{\ta}{\textbf{(1)~}}
\newcommand{\tb}{\textbf{(2)~}}
\newcommand{\tc}{\textbf{(3)~}}
\newcommand{\td}{\textbf{(4)~}}

\definecolor{RebuttalColor}{RGB}{0,0,0}

\DeclareMathOperator*{\argmin}{arg\,min}

\title{Self-supervised perception for tactile skin covered dexterous hands}

\author{
Akash Sharma$^{1,2}$, 
Carolina Higuera$^{1,3}$,
Chaithanya Krishna Bodduluri$^{1}$, \\ \bf
Zixi Liu$^{1}$,
Taosha Fan$^{1}$,
Tess Hellebrekers$^{1}$,
Mike Lambeta$^{1}$, 
Byron Boots$^{3}$, \\ \bf
Michael Kaess$^{2}$, 
Tingfan Wu$^{1}$, 
Francois Robert Hogan$^{1}$,
Mustafa Mukadam$^{1}$\\[3mm]
$^{1}$FAIR at Meta,
$^{2}$Carnegie Mellon University,
$^{3}$University of Washington,
}

\begin{document}
\maketitle

\begin{figure}[htbp]
    \vspace{-5mm}
    \centering
    \includegraphics[width=0.8\linewidth]{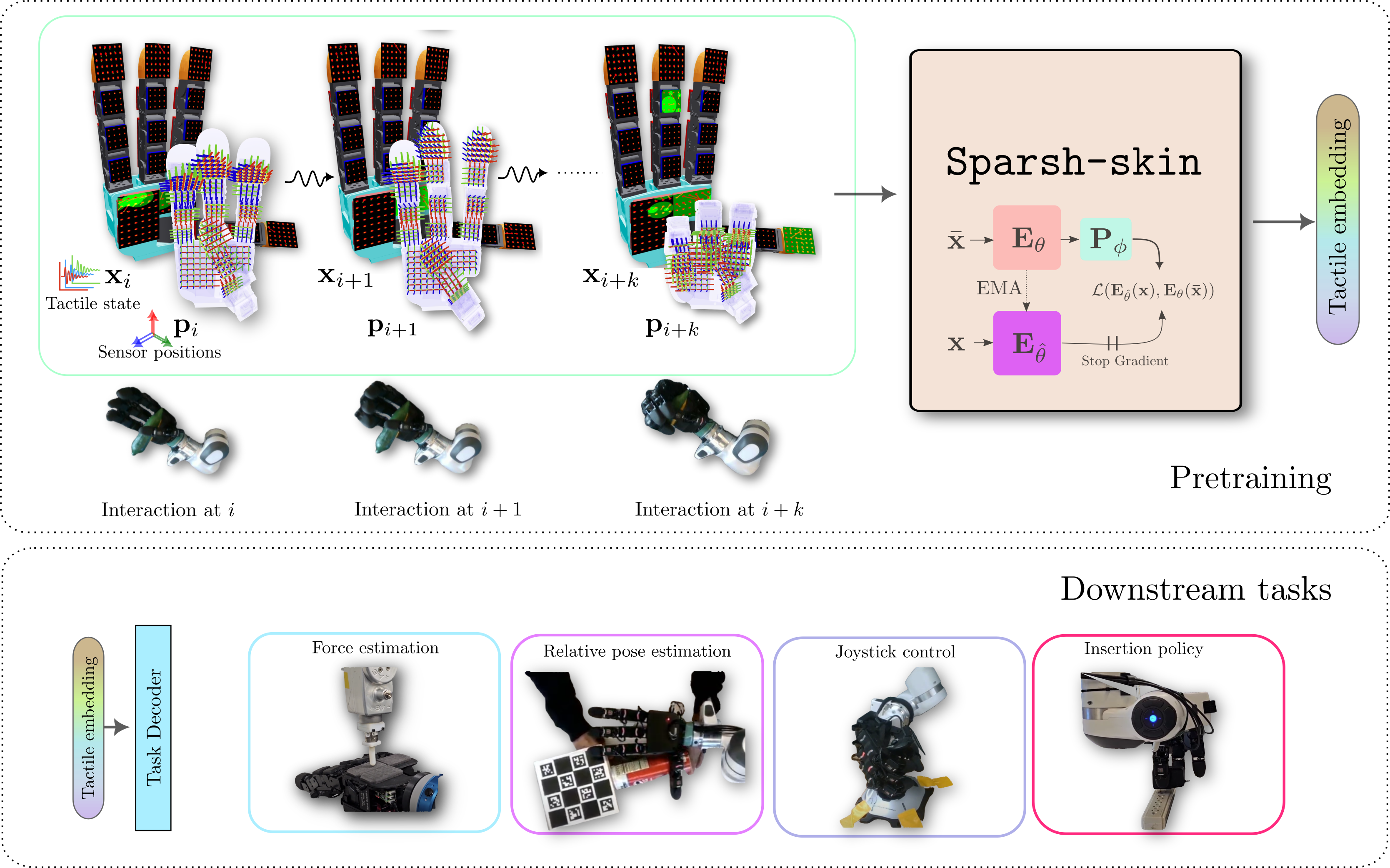}
    \caption{\model is an approach to learn general representations for magnetic tactile skins covering dexterous robot hands. \model is trained via self-supervision on a large pretraining dataset ($\sim 4$ hours) containing diverse atomic in-hand interactions. It takes as input a brief history of tactile observations $\xv_i$ and 3D sensor positions $\pv_i$ to produce performant full-hand contextual representations. \model representations are general purpose and can be used in a variety of contact-rich downstream tasks.\vspace{-2mm}}%
    \vspace{-3mm}
    \label{fig:teaser}
\end{figure}

\begin{abstract}
We present \model, a pre-trained encoder for magnetic skin sensors distributed across the fingertips, phalanges, and palm of a dexterous robot hand. 
Magnetic tactile skins offer a flexible form factor for hand-wide coverage with fast response times, in contrast to vision-based tactile sensors that are restricted to the fingertips and limited by bandwidth. Full hand tactile perception is crucial for robot dexterity. However, a lack of general-purpose models, challenges with interpreting magnetic flux and calibration have limited the adoption of these sensors.
\model, given a history of kinematic and tactile sensing across a hand, outputs a latent tactile embedding that can be used in any downstream task. The encoder is self-supervised via self-distillation on a variety of unlabeled hand-object  interactions using an Allegro hand sensorized with Xela uSkin.
In experiments across several benchmark tasks, from state estimation to policy learning, we find that pretrained \model representations are both sample efficient in learning downstream tasks and improve task performance by over $41\%$ compared to prior work and over $56\%$ compared to end-to-end learning. %
\end{abstract}
\vspace{-3mm}

\section{Introduction}\label{intro}\vspace{-3mm}
Touch is inconspicuous, but plays a crucial role in dexterous manipulation, like when playing the guitar or plugging a cord into a socket when vision is impaired.
The robotics community has leveraged touch to enhance robot learning~\cite{qi2023general, romero2024eyesight, higuera2024sparsh, yang2024anyrotate}, but has so far largely limited their attention to vision-based tactile sensing. %
Recently, sensors such as the DIGIT~\cite{lambeta2020digit}, GelSight~\cite{yuan2017gelsight}, GelSlim~\cite{donlon2018gelslim} and others~\cite{romero2024eyesight, wang2021wedge}, have gained popularity due to their high-resolution output, human-interpretable signals, and accessibility. Capturing touch as images is attractive, as advances in computer vision can be leveraged with minimal friction. Nevertheless, these sensors are slow compared to human skin's touch receptors, come in bulky form factors precluding large area sensing, and are often custom-designed for specific manipulators~\cite{romero2024eyesight}, making reproducibility a challenge. 

Magnetic skin-based sensors such as uSkin (Xela)~\cite{tomo2018xela, tomo2016modular}, ReSkin~\cite{bhirangi2021reskin}, and others~\cite{ bhirangi2024anyskin, PaXini}, offer an alternative for tactile feedback. They provide fast response times ($\sim100$ Hz), lower dimensionality and flexible form factors that can be adapted to complex embodiments, such as multifinger robot hands providing richer states for dexterous manipulation. Despite their potential, the widespread use of these sensors is primarily limited by their complexity: these sensors are difficult to interpret, difficult to model due to hysteresis and other factors, and are primarily hindered by a lack of infrastructure.

Self-supervised learning of general touch representations offers a potential solution: it can learn priors from unlabeled data, making subsequent learning on specific tasks (downstream learning) sample efficient. However, while previous research~\cite{guzey2023dexterity, wu2024canonicalrepresentationforcebasedpretraining}, has applied self-supervision for tactile learning, these approaches often use techniques from computer vision such as treating temporal signals as images~\cite{guzey2023dexterity} and employing masked reconstruction objectives~\cite{wu2024canonicalrepresentationforcebasedpretraining, he2022masked} that may be ill-suited for signals like noisy magnetic flux.

We present \model, a pre-trained tactile encoder model trained using self-supervised learning (SSL) for magnetic skin-like sensors covering %
a multifinger robot hand (see \fig{fig:teaser}). \model directly learns in-hand contact priors from tactile history and hand configuration using a classification objective. Our tactile encoder
simplifies downstream task use, by introducing standardized magnetic time-series data, and reducing the need for real-world labeled data, which is difficult to collect and oftentimes infeasible. For instance, we do not yet have hardware to annotate spatially distributed ground truth force fields. By combining our representation learning algorithm, tactile signal tokenization, and a fully-sensorized multi-fingered hand, we achieve state-of-the-art tactile representations for magnetic-skin sensors, outperforming end-to-end training by $\sim56.37\%$ and prior works by $\sim41.04\%$ on average in both performance and sample efficiency for downstream tasks. We aim to open source our code, datasets, and models. The main contributions of our work are:

\begin{enumerate}[itemsep=-1.0pt,topsep=-2pt,leftmargin=6mm]
    \item \model: a general purpose tactile representation model, trained via self-distillation for magnetic-skin based tactile sensors.
    \item A revisit of tokenization, masking and the learning algorithm choices for temporal magnetic tactile signals which improves downstream task performance by over 41\%.
    \item A dataset containing 4 hours of random play-data of the Allegro robot hand sensorized with the Xela tactile sensors, labeled datasets, metrics, and task design that cover relevant problems in tactile perception to evaluate learned representations.
\end{enumerate}

\vspace{-2mm}
\section{Related work}\label{sec:related}\vspace{-2mm}

\subsection{Tactile sensing}\vspace{-2mm}

Tactile sensors can be broadly categorized into vision-based (e.g. DIGIT~\cite{lambeta2020digit}, GelSight~\cite{yuan2017gelsight}, GelSlim~\cite{donlon2018gelslim} and others~\cite{romero2024eyesight, wang2021wedge}), pressure-based (e.g. force sensitive resistors), impedance-based (e.g. BioTac~\cite{jeremy2012biotac}), and magnetic-based (e.g. uSkin (Xela)~\cite{tomo2018xela, tomo2016modular}, ReSkin~\cite{bhirangi2021reskin}, and others~\cite{ bhirangi2024anyskin, PaXini}) sensors. %
Vision-based sensors commonly used in robot manipulation capture finger-object-environment interactions as images~\cite{lambeta2020digit, yuan2017gelsight}. However, their bulky form factor, low-frequency feedback and high bandwidth requirement limit their application in tasks that require large areas coverage.
Impedance-based tactile sensors offer high temporal resolution, but are difficult to interpret, and currently do not provide full-hand coverage solutions either.
Pressure-based sensors can offer a wide coverage area, but lacks capabilities in shear force sensing.
Magnetic tactile sensors, on the other hand, provide a thin skin-like alternative with options such as ReSkin~\cite{bhirangi2021reskin}, AnySkin~\cite{bhirangi2024anyskin}, and Xela~\cite{tomo2018xela, tomo2016modular} being popular choices. They provide low-dimensional but high-frequency signals. However, when these sensor pads are distributed on all contact interfaces of a robot hand, the total output is high-dimensional. These sensors primarily use hall-effect sensing for force measurement. Xela~\cite{tomo2016modular} in particular works by transducing displacements of permanent magnets embedded in an elastomer arranged in a grid pattern to magnetic flux changes, essentially capturing 3-axis shear and normal forces. ReSkin and AnySkin \cite{bhirangi2021reskin, bhirangi2024anyskin} magnetize the entire elastomer layer continuously, instead of using discrete magnets. This sensing modality has been explored for various contact-rich applications, including planar pushing~\cite{9981270}, surface material classification~\cite{9609678}, grasp stability~\cite{gao2023visuotactilebasedslipdetectionusing}, and policy learning tasks~\cite{10802001, pattabiraman2024learning}.
\begin{wrapfigure}{r}{7.1cm}
    \includegraphics[width=\linewidth]{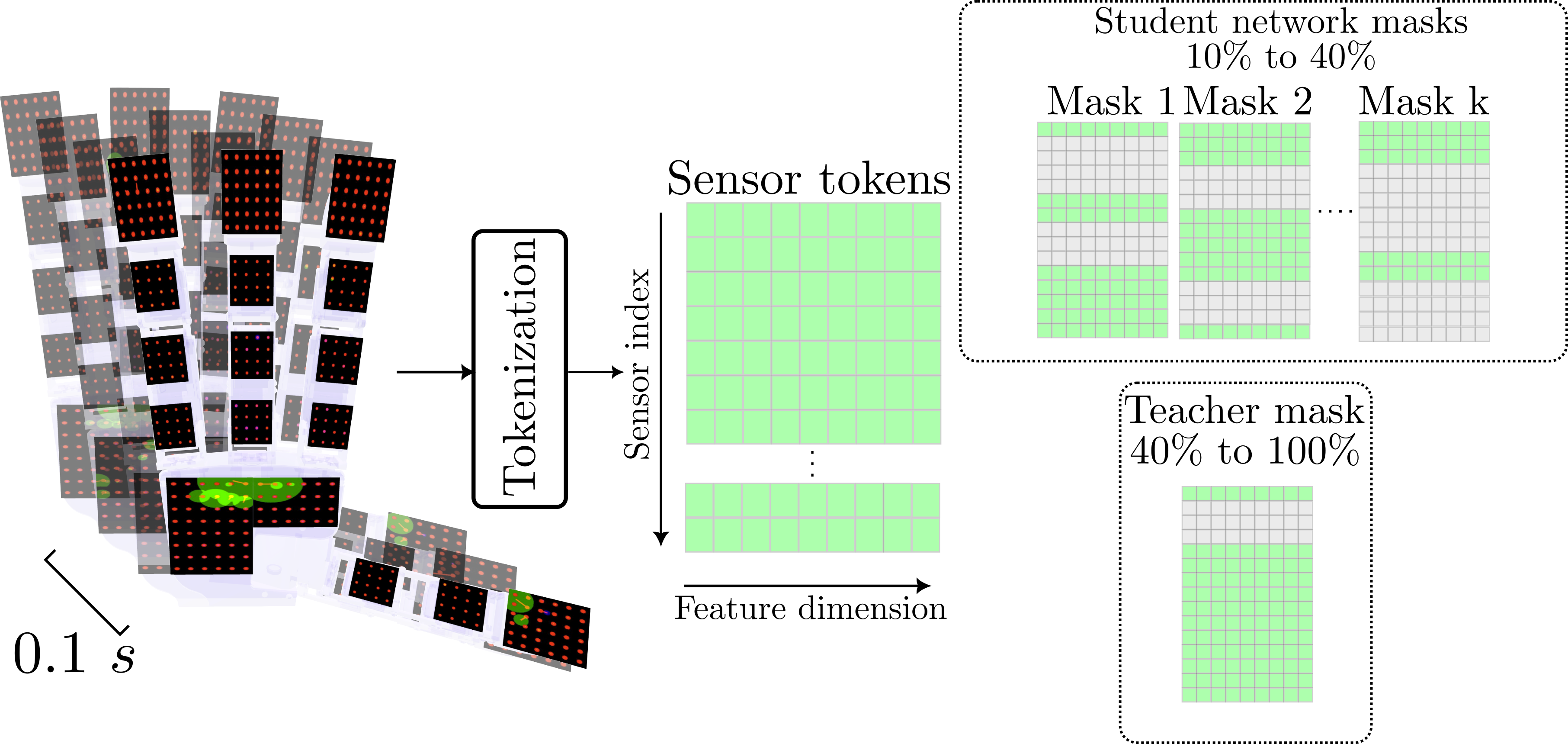}
    \caption{Illustration of Xela signal corruption via masking for SSL prediction task: Once a $100$(ms) window of tactile measurements and sensor positions are tokenized, block masking is applied to corrupt the signal, . For each data sample, the student network receives $k$ different masks, each randomly retaining 10\% to 40\% of the data denoted $\bar{z_i}$. The teacher network, in contrast receives 1-2 masks each retaining 40\% to 100\% of the data denoted $z^*_i$.}
    \label{fig:masking_illustration}
    \vspace{-5mm}
\end{wrapfigure}

\vspace{-3mm}
\subsection{Tactile representation learning}\vspace{-2mm}
Representation learning for vision-based tactile sensors has recently gained significant attention. Since the sensor outputs are images, techniques from computer vision~\cite{he2022masked, van2017neural} have been extended to tactile sensors. This is motivated by a move beyond task-specific encoders to pretrained encoders that promise generalization, with prior work leveraging mask-autoencoders (MAE)~\cite{zhao2024transferable, cao2023learn}, contrastive learning~\cite{yu2024mimictouch, yang2024binding, dave2024multimodal, george2024vitalpretrainingvisuotactilepretraining}, and state-of-the-art methods like self-distillation and joint-embedding predictive architectures~\cite{higuera2024sparsh} to learn tactile representations. %

Research on learning representations for magnetic-based sensors remains relatively underexplored. Since these sensors produce low-dimensional signals, the consensus view is that representation learning is likely unnecessary. However, as we highlight in our work, these signals are indeed high-dimensional due to the complexities of full hand sensing, dynamic tactile signals and hand poses, and magnetic sensor physical properties. This high dimensionality means they benefit from large-scale pretraining to compress information into semantically rich representations that enhance downstream task performance. %
Recently, HyperTaxel~\cite{10802001} applied contrastive learning to learn representations for the Xela sensor for the task of surface recognition but it did not show whether these representations capture contact dynamics. Similarly~\cite{guzey2023dexterity, wu2024canonicalrepresentationforcebasedpretraining} propose representation learning with self-supervised methods such as BYOL~\cite{grill2020bootstrap} and MAE~\cite{he2022masked}. While the idea of representation learning is promising, the choice of meaningful image augmentations without data corruption, is unclear for BYOL. Furthermore, by treating instantaneous tactile measurements as images, these methods discard temporal information and may therefore be suboptimal for tactile tasks that rely on contact dynamics. %
\vspace{-3mm}

\section{\model: self-supervised representations for tactile skins}\vspace{-2mm}
\model is a self-supervised modeling approach to learn from random-play data, generalizable tactile features for dexterous hands equipped with magnetic-skin tactile sensors. 
\vspace{-2mm}
\subsection{Preliminaries}
\paragraph{Self-distillation for representation learning} Self-distillation \cite{oquab2023dinov2, baevski2022data2vec, grill2020bootstrap} is a powerful paradigm in self-supervised learning involving a pair of identical neural networks, termed the student $\Ev_\theta$ and teacher network $\Ev_{\hat{\theta}}$. The student network receives a corrupted version of a data signal $\bar{\xv}$ that is to be encoded, while the teacher network receives privileged information about the same data sample $\xv$. Then, the student network is tasked with predicting through a small predictor network $\Pv_\phi$, the data representation that the teacher produced. To prevent the teacher from producing degenerate representations -- for instance, a constant representation for all data -- the teacher weights are not updated via back-propagation, but only through an exponential moving average (EMA) of the student weights. Specifically, the following objective is optimized: 
\begin{align}
    \argmin_{\theta, \phi} \norm{\Pv_\phi(\Ev_\theta (\bar{x})) - \text{sg}(\Ev_{\hat{\theta}} (\xv))}
    \vspace{-2mm}
\end{align}
where $\text{sg}$ indicates stop gradient, and $\hat{\theta} \triangleq \text{EMA}(\theta)$.
Since the teacher network is an exponential moving average (EMA) of the student work, knowledge is \emph{self-distilled} through the representation prediction task. 
\vspace{-3mm}
\paragraph{Robot setup and pretraining data}\label{subsec:setup} Our setup consists of the Allegro hand sensorized with Xela uSkin, attached to a Franka Panda robot arm. The Allegro hand is equipped with 18 Xela uSkin sensing pads, consisting of 4 curved fingertip sensors with 30 individual sensors, 11 4x4 grid sensors pads attached to the finger phalanges, and 3 4x6 sensing pads attached to the palm, resulting in a total of 368 individual sensors. %
\begin{wrapfigure}{r}{7.1cm}
        \centering
        \includegraphics[width=\linewidth]{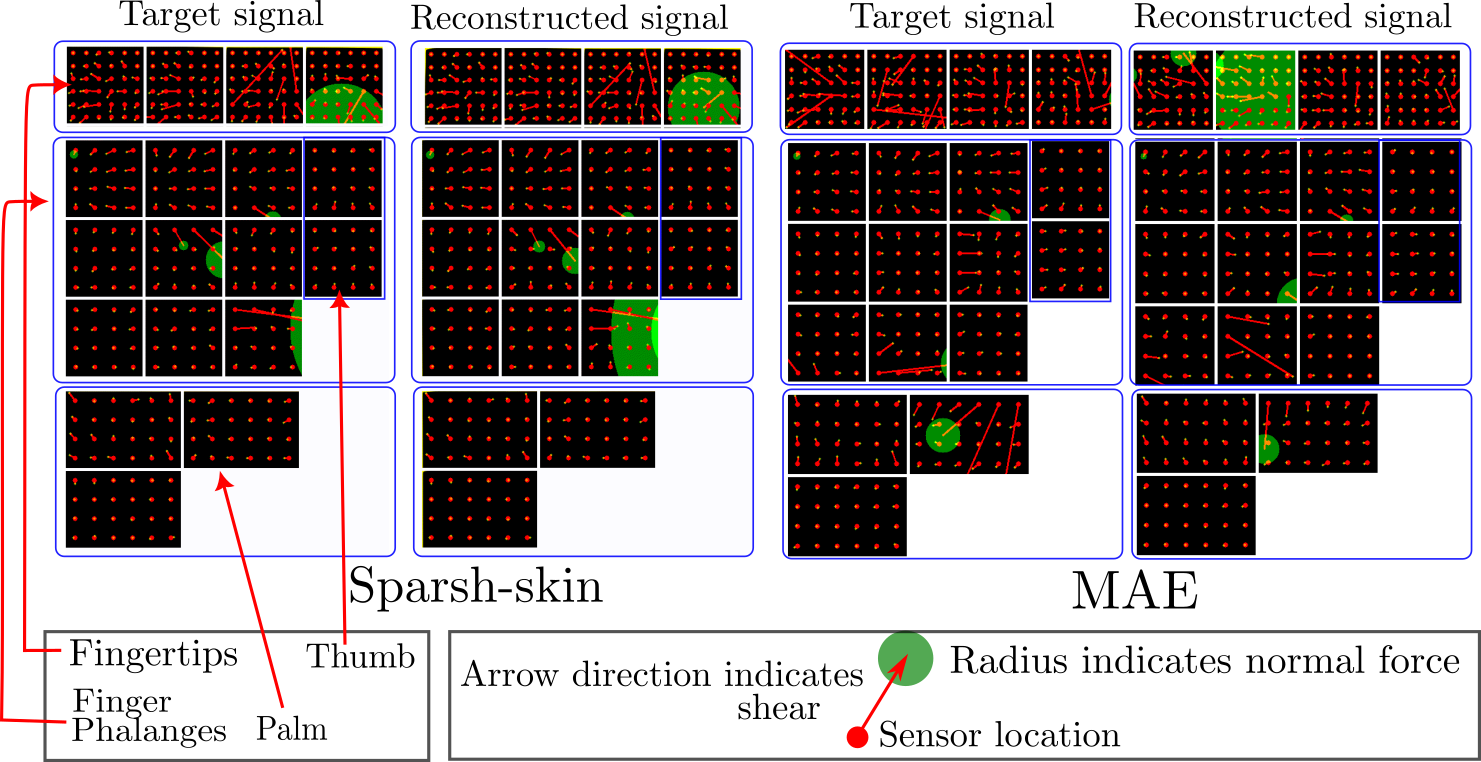}
    \caption{\small{Visualization of reconstructions from the reconstruction online probe. When compared to MAE, \model reconstructs signals effectively. Specifically, note that the normal forces and directions are better preserved by \model. Here, we visualize a single frame from a 0.1s tactile window.}} %
    \label{fig:reconstruction_example}
    \vspace{-5mm}
\end{wrapfigure}
We collected a dataset of the hand performing various atomic manipulation actions with 14 household objects and toys, including squeeze, slide, rotation, pick-and-drop, circumduction, pressing, wiping, and articulation. Using a VR-based teleoperation system with Meta Quest 3, which builds on the inverse kinematics-based re-targeting method proposed in~\cite{handa2020dexpilot}, we recorded 11 sequences (approximately 2 minutes each) for each object, totaling around 4 hours of varied interactions. The dataset includes top/left camera views, Xela signals, and robot and hand joint states, covering a range of rigid and deformable objects with diverse tactile properties (see \fig{fig:umap_vis}). %

\vspace{-1mm}
\subsection{Architecture}
\vspace{-2mm}
\model uses a Transformer~\cite{vaswani2017attention} as the student and teacher network for self-distillation. 
\vspace{-3mm}\paragraph{Sensor tokenization} We perform baseline subtraction on Xela signals to account for their uncalibrated nature and consistent biases. A single baseline signal is collected with the Allegro hand in a resting configuration (palm up and flat) and used for all downstream tasks, unlike prior work~\cite{bhirangi2021reskin, pattabiraman2024learning}, which collects a new baseline signal per training sequence. We also resample Xela signals to a consistent 100Hz frequency. as the sensor data rate fluctuates between 80Hz to 100Hz, unlike prior work \cite{guzey2023dexterity} that subsamples data to match modalities at lower frequencies.

We note that for representation learning, tactile data can be temporally correlated, and instantaneous signals cannot provide context for contact changes, therefore we choose to learn representations for chunks of $100$ms of data. First, inputs to \model are formatted corresponding to a brief history of $0.1$ seconds of the sensor signal ${x_{1:10} \in \mathbb{R}^{10 \times 368 \times 3}}$ concatenated with the history of sensor position ${p_{1:10}\in \mathbb{R}^{10 \times 368 \times 3}}$ computed from the forward kinematics of the Allegro hand. Inputs are then tokenized through a linear projection $f_\text{linear}$ to the dimension $d$ of the representation ${z_i=f_{\text{linear}}(x_{1:10} | p_{1:10}) \in \mathbb{R}^{368 \times d}}$. Finally, a learnable token is added to each sensor according to the three types of Xela sensing pads (see \ref{subsec:setup}) on the Allegro hand. We do not add additional positional embedding and instead rely on the sensor position to provide 3D positional information to the transformer network.

\vspace{-2mm}\paragraph{SSL prediction task} Although cropping and resizing images is a common technique for signal corruption in the image domain, applying this method to magnetic flux readings alters the shear profile. %
Therefore to avoid any untoward data augmentation that changes the semantic meaning of the signal, we use block masking \cite{assran2023self} to corrupt signals that are input to the encoding networks. Specifically, input data is masked after sensor tokenization in a cross-taxel manner i.e., given tokenized data from 368 sensors, we mask sensor data from local contiguous blocks including sensors even from neighboring sensor islands by removing those sensors from the input (see \fig{fig:masking_illustration}).  %
The masked sensor tokens are subsequently transformed through the student and teacher network as $\Ev_\theta(\bar{z_i})$ and $\Ev_{\hat{\theta}}(z^*_i)$ respectively.

For the prediction task, we use classification by defining a set of prototype classes as in~\cite{caron2021emerging, oquab2023dinov2}, which is robust to sensor noise compared to masked auto-reconstruction. The sensor tokens after transformation are  converted into prototype logits through a classification head $f_\text{class}$ as $\bar{p}_i$ and $p^*_i$ respectively for the student and teacher networks. We use both the class token and the patch level cross entropy objective between the student and teacher logit predictions to enforce local-to-global correspondence learning in the sensor representation. Additional details about the model architecture, MAE reconstruction comparisons and training hyper-parameters are in the Appendix.

\begin{wrapfigure}{l}{4.0cm}
    \centering
    \includegraphics[width=\linewidth]{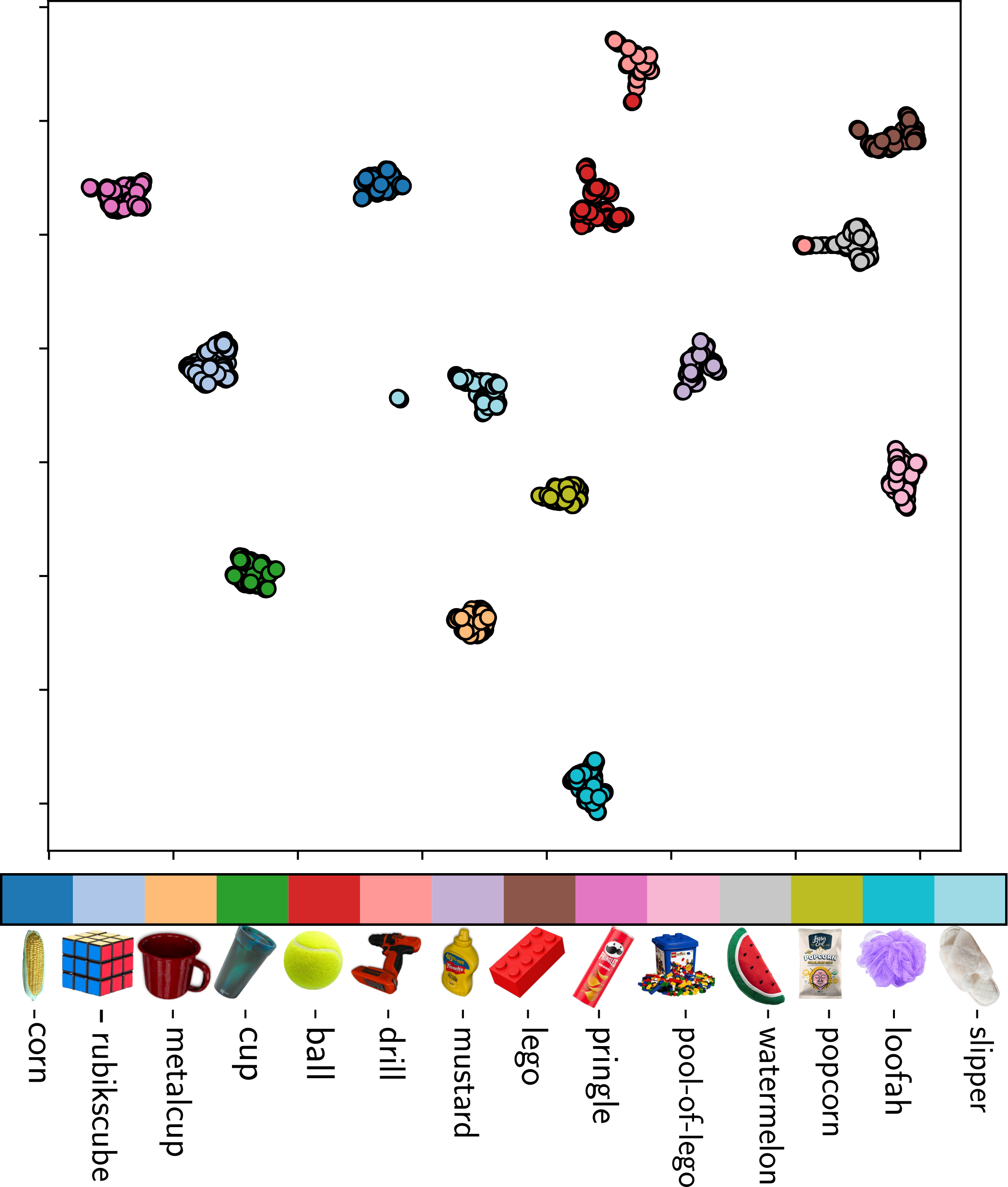}
    \caption{\small{UMAP visualization of representations colored by object in robot hand.\vspace{-2mm}}} %
    \label{fig:umap_vis}
    \vspace{-2pt}
\end{wrapfigure}
\vspace{-2mm}
\paragraph{Online Probes} Unlike supervised learning (SL), where model performance is easily monitored through training and validation losses, in self-supervised learning (SSL), prediction task losses do not directly convey downstream task performance. In fact, in the presence of an EMA teacher network, which acts as a moving target, the prediction task loss can increase in tandem with the predictions of the teacher network. Therefore, we rely on online probes to monitor downstream performance. During training, we evaluate the tactile representation for a) reconstruction and b) the ability to identify objects used in play data.

\fig{fig:reconstruction_example} provides a qualitative visualization of the reconstruction performance obtained by the decoder using representations computed by the student network $\Ev_\theta(\bar{z})$. Here, we find that \model trained using the MAE reconstruction objective (identical tokenization) scheme, is significantly inferior at reconstruction compared to \model trained via self-distillation.
In terms of object classification performance, we achieve approximately 95\% accuracy across 14 classes, while both BYOL (treating tactile signal as images)~\cite{guzey2023dexterity} and MAE (using tactile and proprioception history)~\cite{he2022masked} are limited to $\sim81\%$ accuracy. Additionally, \fig{fig:umap_vis} presents a UMAP~\cite{mcinnes2018umap} visualization of the representations, where sequences from each object are mapped to distinct, non-overlapping clusters. 
\vspace{-2mm}\paragraph{Implementation Details}
Our method is designed for the Xela sensor but can be extended to any skin sensor with 3-axis time-series output signals. \model is trained for 500 epochs on 8 Nvidia A100 GPUs with a batch size of 64, using the AdamW optimizer, and linear warmup followed by cosine schedule as the learning rate scheduler. Downstream tasks are trained with task-labeled data on 1 Nvidia A100 / 4090 GPU. Furthermore, \model supports realtime inference with an inference time of $\sim$7ms. Please refer to the appendix for additional training details. 

\vspace{-3mm}\section{Experiments}\vspace{-2mm}
In this section, we assess the ability of \model to comprehend tactile properties, enhance perception, and enable policy learning for manipulation  %
through four downstream tasks spanning tasks studied in the tactile sensing literature: namely \ta Force estimation, \tb Joystick state estimation, \tc Pose estimation, and \td Policy learning via the plug insertion task.
\begin{figure}[htbp]
    \begin{subfigure}[t]{0.49\linewidth}
    \centering
    \includegraphics[width=\linewidth]{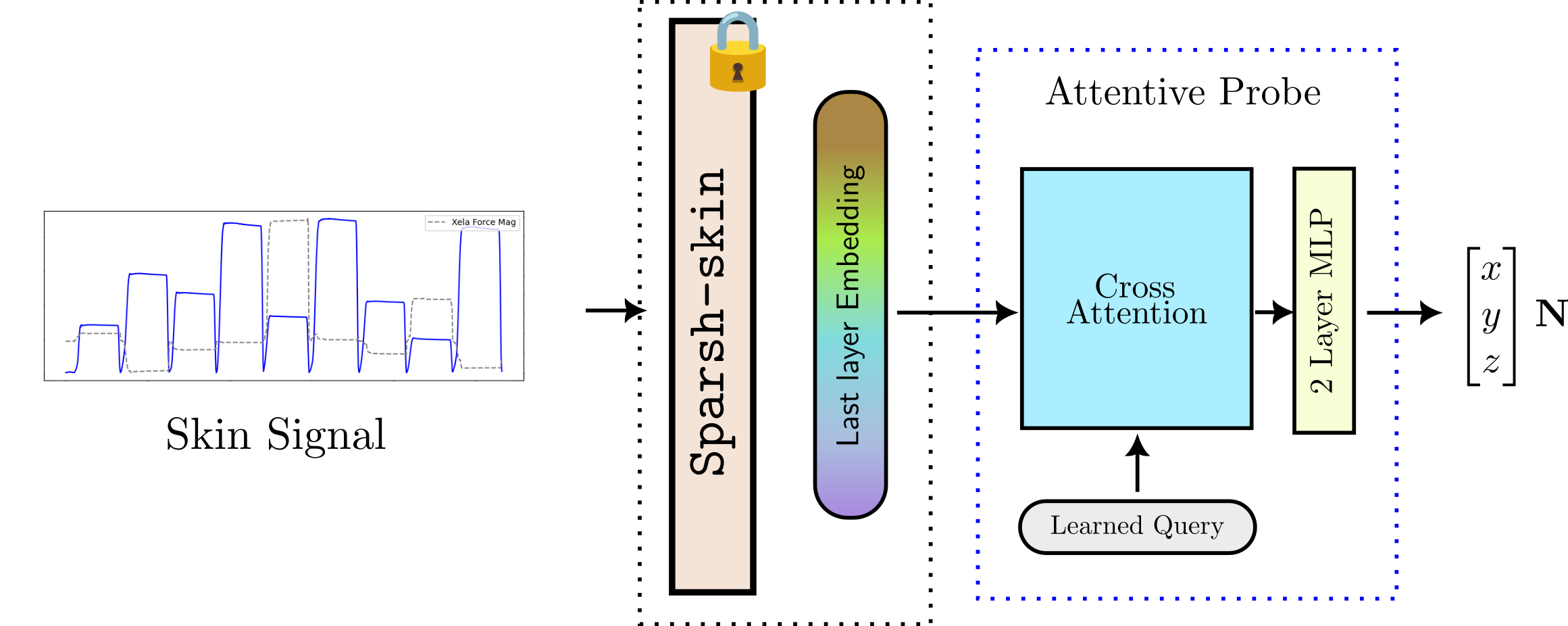}    
    \caption{Attentive probe: Attentive pooling + small 2-layer MLP for regression tasks}
    \label{fig:decoder_arch_simple}
    \vspace{2mm}
    \end{subfigure}
    \begin{subfigure}[t]{0.49\linewidth}
    \centering
    \includegraphics[width=0.7\linewidth]{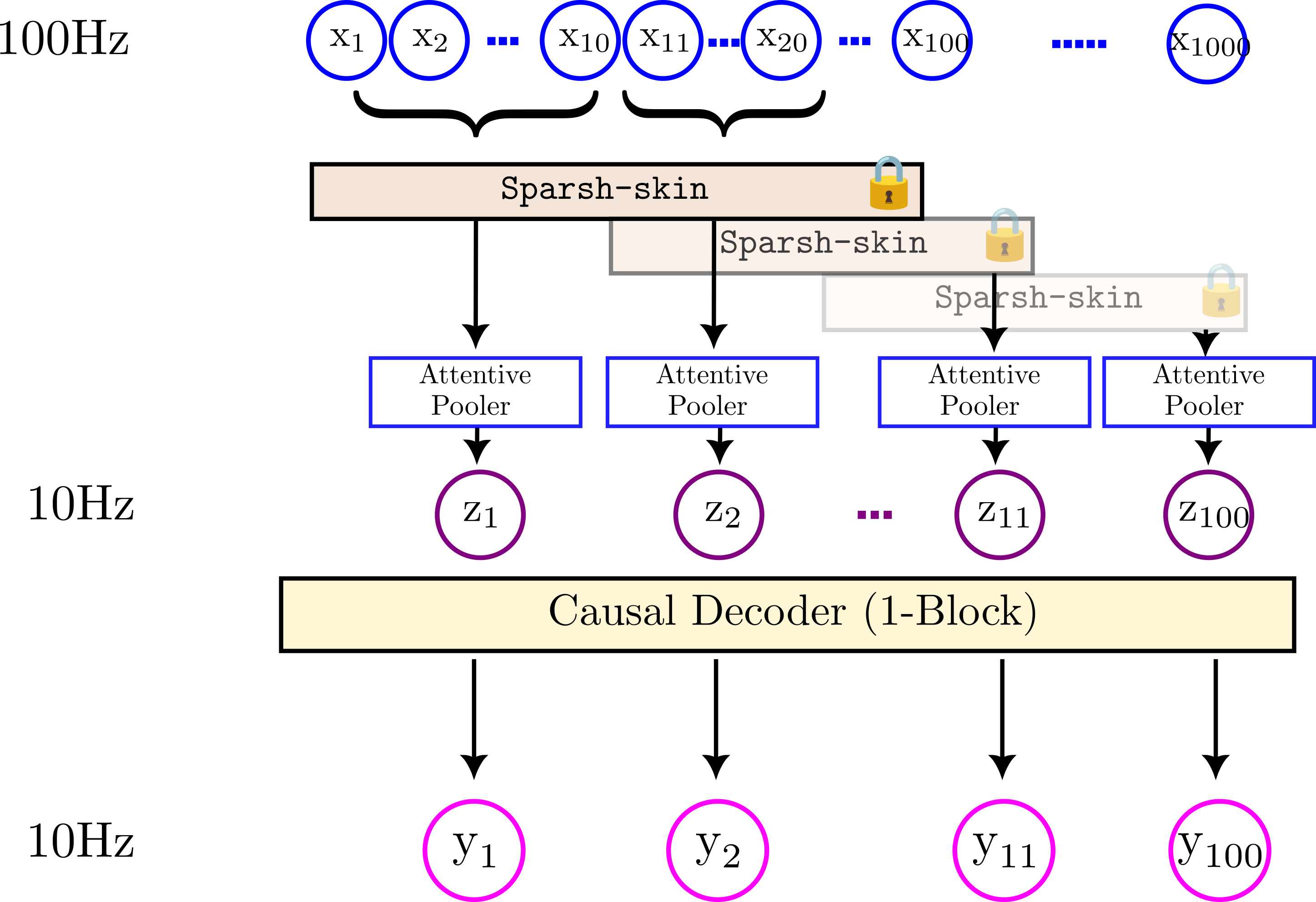}
    \caption{Decoder with a 1-layer transformer block for sequence to sequence prediction tasks}
    \label{fig:decoder_arch_sequence}
    \end{subfigure}
    \caption{We use two types of decoders for (a) instantaneous, and (b) temporal tasks. Both decoders contain the attentive pooler which uses a learned query token to cross attend to sensor features to output a \emph{single token full-hand} representation.}
    \vspace{-2pt}
\end{figure}
\vspace{-2mm}
\subsection{Evaluation protocol}\vspace{-2mm}
\paragraph{Downstream task decoders.}
The tasks we consider are of two types: a) requiring instantaneous prediction, and b) requiring temporal reasoning over tactile data. For tasks such as force estimation that require an instantaneous estimate, we use attentive pooling (see~\fig{fig:decoder_arch_simple}). For tasks such as pose estimation and joystick state estimation, that require sequence reasoning, tactile observations are transformed into tokens at the output frequency through a cascaded application of the backbone network. This is followed by attentive pooling as illustrated in~\fig{fig:decoder_arch_sequence}.%
\vspace{-3mm}\paragraph{Model comparisons.} For each of the downstream tasks, we explore multiple variants of the \model encoder, along with additional baselines:
\begin{enumerate}[itemsep=-1.5pt,topsep=-2pt,leftmargin=5mm]
    \item BYOL$^*$, our reproduction of the BYOL~\cite{grill2020bootstrap} approach to tactile representation learning following~\cite{guzey2023dexterity} using our collected play data and tactile data formatted as images, since the setup used in~\cite{guzey2023dexterity} does not contain palm sensing and uses an older variant of the tactile sensor.
    \item End-to-end, training the entire encoder-decoder network with same capacity as \model using only labeled task data
    \item \model (frozen), pretrained representation that uses tactile and hand configuration history.
    \item \model (finetuned), where the encoder network is finetuned with task-specific data.
    \item \model (MAE), pretrained representation that uses tactile and hand configuration history trained using MAE supervision instead of self-distillation.
\end{enumerate}
For tasks \textbf{(1 - 3)}, we measure performance using the average root mean squared error (RMSE). Additionally, we evaluate each method for sample efficiency by reducing the downstream labeled data accessible during training. Then, for \td plug insertion, we measure success rate (SR) across trials where we select the best model from \textbf{(1 - 3)} for comparison against the end-to-end baseline. %

\begin{figure}[!t]
    
    \centering
    \includegraphics[width=0.8\linewidth]{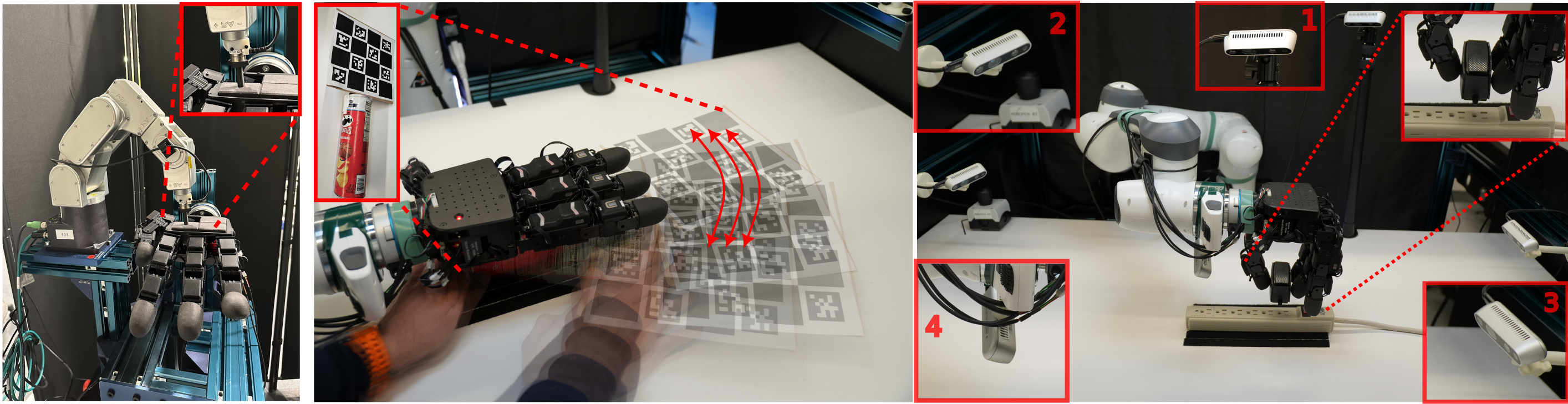}
    \caption{Hardware setup used for downstream tasks: \textbf{(Left)} shows the setup for force estimation. We use 3D printed probes attached to a F/T sensor to indent onto the Xela sensors. \textbf{(Middle)} shows the setup for pose estimation. We track an object mounted with an ArUco marker to obtain ground truth pose estimates while randomly moving it under the robot hand. \textbf{(Right)} shows the setup for plug insertion policy task. We collect tactile measurements and camera observations from three third-person view cameras and a wrist camera view.\vspace{-10pt}}
    \label{fig:task_setup}
    \vspace{-2mm}
\end{figure}
\vspace{-3mm}\subsection{Downstream tasks}
\begin{figure}
    \centering
    \includegraphics[width=\linewidth]{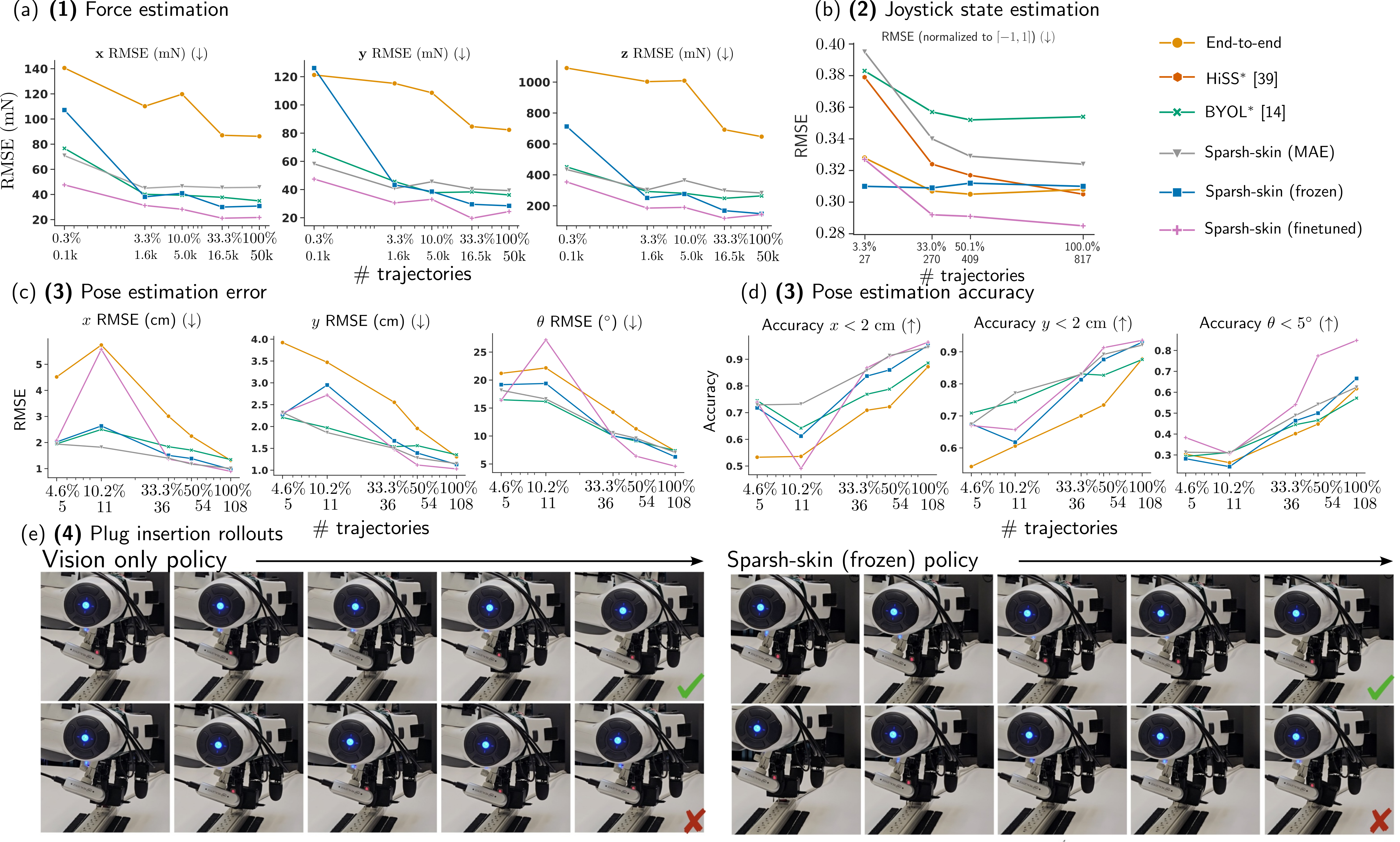}
    \caption{Summary of results comparing \model on all tasks. \textbf{(a)} Force estimation (RMSE ($\downarrow$)): BYOL pre-training is less accurate at predicting normal forces. \textbf{(b)} Joystick state estimation ($\downarrow$): \model outperforms end-to-end overall and is competitive with HiSS$^*$ even when it is given access to only 3.3\% of dataset. \textbf{(c)} Pose estimation error ($\downarrow$) and \textbf{(d)} Pose estimation accuracy ($\uparrow$): \model (finetuned) has a $\sim 10$\% improvement over end-to-end for translation and $\sim 20$\% improvement for rotation. (e) Snapshots of plug insertion policy rollouts (success and failure). Vision-only policy succeeds primarily when the starting position is directly above the socket, while \model (frozen) achieves 75\% success rate, with failures mainly due to loss of grip when sliding to locate the socket\vspace{-5mm}}
    \label{fig:percepskin_plot_summary}
\end{figure}

\vspace{-2pt}
\paragraph{\ta Force Estimation.} This task involves regressing tactile signals to 3-axis normal and shear forces on a robot hand's palm. We collected force-labeled data using a robot arm with an F/T probe to apply varying normal forces (0.25-5.0N) with hemispherical and flat indenters (see~\fig{fig:task_setup}~(left)). The probe's position was randomly sampled across the sensor pad, including locations both on and between magnetometers, differing from sensor characterization which only tests atop magnetometers.
\textbf{Results} (see \fig{fig:percepskin_plot_summary}(a)) %
While the end-to-end model is particularly worse at predicting forces throughout the spectrum, in low data regimes -- 3.3\% to 10\% of the labeled data in this case -- it is interesting to note that \model (finetuned) and \model (frozen) do not see any significant loss in performance. To this end, we test the models with even smaller number of downstream task data samples to find that \model (finetuned) is able to predict forces at a reasonable accuracy (350 mN in $z$) even with only $\sim 100$ samples. Additionally,  we find \model (MAE) is  worse at predicting forces highlighting that MAE may not be suitable for noisy magnetic flux signals. Furthermore, BYOL$^*$ is competitive albeit marginally inferior with respect to \model (frozen) as this task tests for instantaneous force decoding. Additional details and results are in the appendix. 
\vspace{-4mm}\paragraph{\tb Joystick state estimation.}  We adapt this task from~\cite{bhirangi2024hierarchical} (see \fig{fig:teaser}), as a study of \emph{full-hand} object state estimation. The task is a sequential problem of predicting the joystick states (roll, pitch and yaw) given a short tactile history. %
In addition to the comparison of \model with an end-to-end approach, we also compute the RMSE results from the best reported model in HiSS~\cite{bhirangi2024hierarchical} (denoted as HiSS$^*$ in \fig{fig:percepskin_plot_summary}(b)). Additional data and pre-processing details are in the appendix.%

\textbf{Results} Our model (\model) matches baseline (HiSS~\cite{bhirangi2024hierarchical}) performance using full data, despite challenges from jittery teleoperation such as inconsistent touch even with similar joystick maneuvers. Notably, \model (frozen) achieves similar performance even with only 3.3\% of the data, demonstrating high sample efficiency. \model consistently shows lower prediction error across data budgets (\fig{fig:percepskin_plot_summary}(b)). Furthermore, \model (finetuned) drastically speeds up training, reaching comparable performance to an end-to-end approach in 12k optimization steps versus 220k (a 95\% speedup) when using a 33\% data budget. An illustration is provided in the Appendix.

\vspace{-3mm}\paragraph{\tc Pose estimation.} %
This task tests the ability to track and accumulate slip under the sensors to predict object pose changes  $(\mathbf{t}_i^R \triangleq (x,y, \theta)) \in \mathbf{SE(2)}$ using the setup in \fig{fig:task_setup}~\textcolor{black}{(middle)}. %
We collect 120 trajectories ($\sim30$s each), by manually sliding/rotating an object in a range of $\sim(25\text{cm}, 25\text{cm}, 100^\circ)$ under the Allegro hand, tracking ground truth object pose using ArUco tags. These poses in the camera frame are %
transformed into the robot hand frame and then projected into $\mathbf{SE}(2)$. %
We use the sequence decoder (~\fig{fig:decoder_arch_sequence}) which processes 1-second windows of tactile data (100Hz) and object pose (10Hz). In addition to RMSE, for this task, we also measure performance via pose accuracy (proportion of predictions within 2cm translation and 5° rotation error).%

\textbf{Results} (see \fig{fig:percepskin_plot_summary}(c)(d)) All representations models pre-trained on play data achieve lower RMSE and higher pose prediction accuracy than the traditional end-to-end approach. In particular for \model (finetuned) we find a $\sim 10\%$ improvement over the end-to-end model with the full dataset for translation, and $\sim 20\%$ improvement for rotation. In low data regimes (33\%) \model (MAE) outperforms other models since there is a direct correlation between translation and the displacement of the magnetometers on the Xela sensors, while BYOL$^*$ shows similar performance to \model (frozen) which maintains $\sim$70\% accuracy. %
Additionally, we observe that allowing in-domain data to fine-tune the \model representations is advantageous, especially for better tracking rotation of the object, which is harder, as it involves torsion.

\begin{wraptable}{r}{6cm}
\centering
\vspace{-10pt}
\begin{tabular}{@{}lc@{}}
\toprule
Model                   & SR ($\uparrow$) \\ \midrule
VisuoSkin~\cite{pattabiraman2024learning}& $0.66$  \\
Vision only (V)             & 0.20 \\ 
End-to-end V+T              & 0.40 \\            
\model V+T (frozen)         &\cellcolor{tabfirst}0.75 \\
\model V+T (finetuned)      &\cellcolor{tabsecond}0.70 
\end{tabular}
\caption{Policy learning (plug insertion): Success rate percentage reported over 20 trials, while ensuring identical initial conditions during each trial for the tested policy variants. VisuoSkin results are obtained from \cite{pattabiraman2024learning}}
\label{tab:t4_sr_plug_insertion}
\vspace{-1em}
\end{wraptable}

\vspace{-3mm}\paragraph{\td Policy learning (plug insertion).} We train a transformer decoder policy predicting action chunks~\cite{zhao2023act} with \model representations as input for this task. We adapt the insertion task ~\cite{dong2021tactile, wu2024tacdiffusion, pattabiraman2024learning} as it is fundamentally tactile requiring touch feedback to observe the alignment state of the plug. The task involves inserting a pre-grasped plug into a fixed socket using a 7-DOF Franka arm and Allegro hand (Fig.~\fig{fig:task_setup} right) unlike~\cite{pattabiraman2024learning} which used parallel jaw grippers. We collected 100 demonstrations via kinesthetic teleoperation, recording synchronized data: four camera views $(\Iv^\text{left}_t, \dots)$, Allegro tactile readings $(z_t)$, and robot joint states. The arm's initial position was randomized within a $0.05m\times0.05m\times0.02m$ volume $\sim10$cm above the socket, while the socket position is fixed. The policy predicts sequences of absolute end-effector poses (3D position + axis-angle orientation) $\av \triangleq (\Tv_t, \Tv_{t+1}, \dots)$, conditioned on visual and tactile observations but not proprioception (joint states). We evaluate average success rate over 20 trials with randomized start positions, comparing \model variants (V + \model (frozen), V + end-to-end, V + \model (finetuned)) against a vision-only (V) baseline to assess tactile contribution. Further details are in the Appendix.

\textbf{Results}: (see Table~\ref{tab:t4_sr_plug_insertion}) We find that policies conditioned on pretrained \model features outperform the end-to-end model. In~\fig{fig:percepskin_plot_summary}(e) (see supplementary for video), we present snapshots of real-world policy deployments for both vision-only and visuo-tactile \model (frozen) policies.%
Without tactile modality (vision only), we find that the policy is able to get close to the socket but indefinitely continues to search for it and does not push the plug in, even when it is directly above the socket. Further, we find that this policy tends to keep pushing the plug to the left of the socket, which we hypothesize is due to perceptual aliasing, where the plug incorrectly appears to be right above the socket from the wrist camera. On the other hand, all model variants with access to the tactile modality observe respectable success rates. In qualitative inspection, we find that the policies using \model (V+T frozen) representations slides after making contact with the extension board, while \model (V+T end-to-end) and \model (V+T finetuned) tends to retry by lifting the plug, when mistakes occur. As noted earlier, in comparison with \cite{pattabiraman2024learning, bhirangi2024anyskin} which trains end-to-end visuo-tactile policies, our setup uses a multifinger Allegro hand as the manipulator, where the plug is grasped using three fingers; nevertheless, we find that policies trained with tactile features from \model are competitive. 
\vspace{-2pt}

\vspace{-3mm}\section{Conclusion}\vspace{-3mm}
We present \model, a high-performance tactile representation model trained via self supervision for magnetic skins on dexterous hands. %
Through evaluations across tactile-centric estimation and policy learning tasks, 
we demonstrate the efficacy of our supervision objective, tokenization, masking strategies and pretraining of the model over a large unlabeled dataset containing $\sim$4 hours of atomic contact interactions with household objects. In experiments, when considering sample efficiency (training on 33\% downstream data), we find that \model (frozen) outperforms the end-to-end model baseline by $\sim$56\%, our adaptation of the BYOL approach to tactile representation learning by $\sim$28\% and the \model (MAE) baseline by $\sim$53\%. We believe that \model represents a step toward foundation models for full-hand tactile representations that enables high-dexterity robotics tasks. 
\newpage

\section*{Limitations.} 
We identify the following limitations for the model design and evaluation framework:
\begin{enumerate}
    \item \textbf{Model design:} %
    Although \model handles contact dynamics implicitly by learning representations over windows of tactile signal, the data corruption strategy is inherently spatial. Future work may consider explicitly handling temporal correlation learning via temporal prediction tasks.
    
    \item \textbf{Pose estimation:} Our pose estimation task is not designed for applications where the hand is not static. While pose estimation in its current iteration is designed to track a 2D pose with a flat fixed hand configuration, real object interactions involve both 3D pose changes, and simultaneously changing hand configurations. There are also exciting avenues to study tasks such as grasp stability prediction~\cite{roberto2018graspstability} and slip prediction.
    
    \item \textbf{Manipulation policy:} Although our experiments with the real physical system support the hypothesis that tactile information from magnetic skin sensors improves policy performance, we still need to assess their ability to generalize. Visuo-tactile policies can overfit to the specific tactile signatures of objects and environments used in data collection, which raises an open question: how can we achieve generalization across diverse tactile feedback signals while maintaining data efficiency?
    
\end{enumerate}

\acknowledgments{The authors thank Unnat Jain, Tarasha Khurana, Hung-Jui Huang, Jessica Yin, Changhao Wang, Luis Pineda, Mrinal Kalakrishnan and Youngsun Wi for helpful discussion and reviews of the paper. This work is supported by Meta FAIR labs.}

\bibliography{references}  %

\newpage
\appendix
\section*{Appendix} \label{appendix}
\begin{figure*}[htbp]
    \centering
    \includegraphics[width=\linewidth]{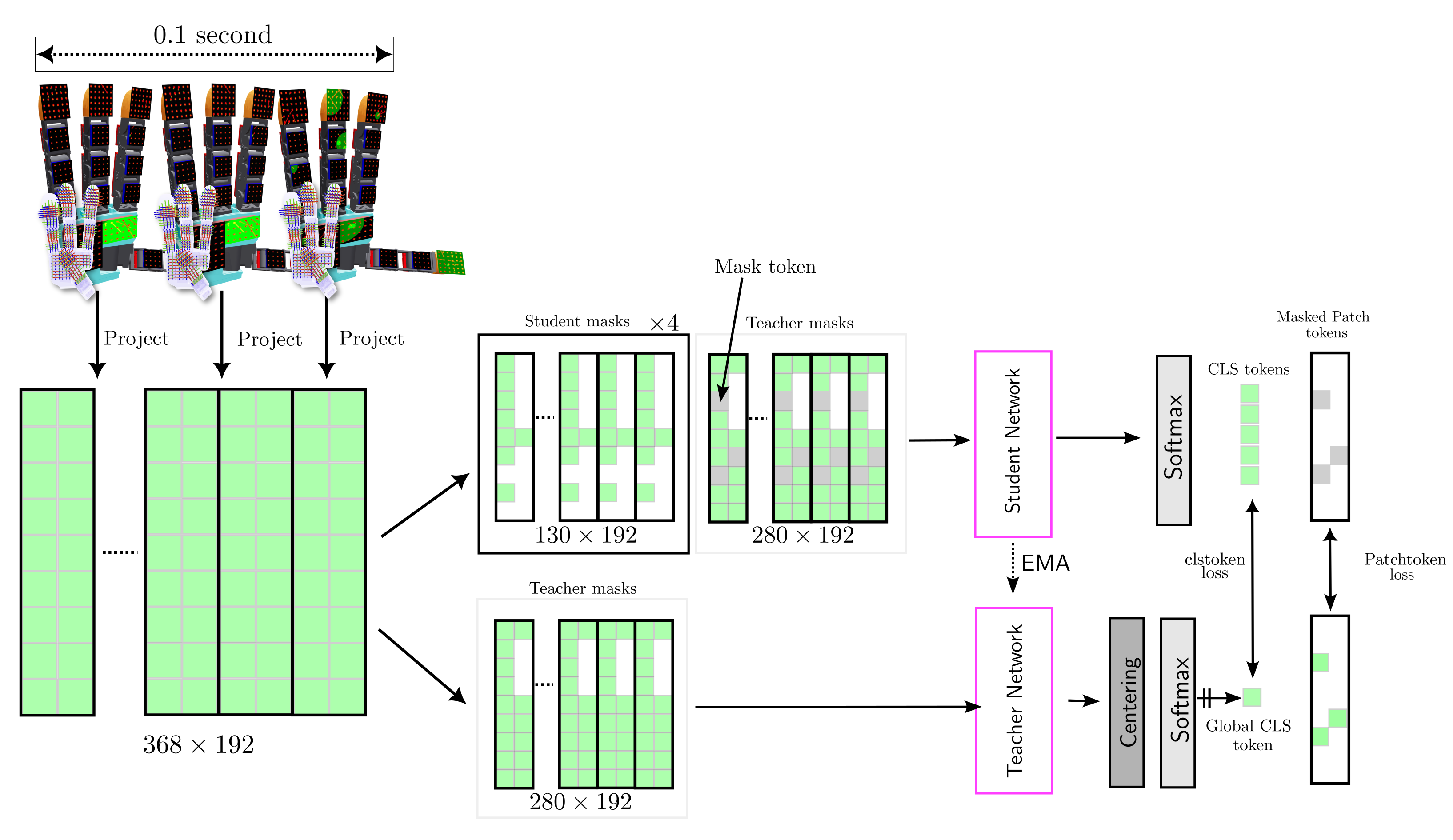}
    \caption{\textbf{\model block diagram for self-supervised learning of skin representations.} Our approach follows the student-teacher framework and loss functions used in self-distillation. However, we adapt the transformer input tokenization to accommodate time-series Xela data.}
\label{fig:xela_architecture}
\end{figure*}

\section{\model self-supervision details} \label{appendix_architecture}

\subsection{Training details}
We train \model on 8 Nvidia A-100 (80G) GPUs. To monitor learning, we use reconstruction online probe and classification via linear probing. We use AdamW optimizer and use a linear rampup followed by a cosine schedule as the learning scheduler. Further, we find that tuning momentum value as well as the weight decay factor was important in observing training convergence. Additional information of hyperparameters is detailed in Table ~\ref{tab:appendix-hparams}.

\begin{table}[!h]
\centering
\begin{tabular}{lc}
\toprule
Architecture & ViT-Tiny (adapted) \\
Embedding dim & 192 \\
EMA decay & [0.994, 1.0] \\
LR & 1e-4 \\
Batch size & 64 \\
\bottomrule
\end{tabular}
\caption{
\textbf{Training hyperparameters for \model.}
All models run for 500 epochs with optimizer AdamW, a weight decay cosine schedule from 0.04 to 0.4, and a  learning rate warmup of 30 epochs.).
}
\label{tab:appendix-hparams}
\end{table}

\subsection{Architecture details}
Our encoder model is a modified version of Vision Transformers~\cite{dosovitskiy2020image}. Specifically, we adapt the tokenization of the time-series Xela with sensor pose data. After flattening the 3D-axis magnetic reading per magnetometer (368) and concatenating their corresponding pose in chunks of 0.1 second, the inputs $x \in \mathbb{R}^{10 \times 368 \times 6}$ are tokenized through a linear projection to the dimension $d$ of the representation $f_{linear}(x) \in \mathbb{R}^{368 \times d}$. We use a tiny model with $d=192$. We add a learnable embedding to identify different types of xela pads (palm, phalanges and fingertips). Then, we construct different cropped view of the data, two global views and eight local views. We mask sensor data from contiguous blocks by removing those sensors from the input. For the local view we retain between 10\% and 40\% of the tactile signal, whereas for the global views we retain 40\% to 100\%. An illustration of the masking and diagram block of the pipeline for self-supervised learning of Xela representations is shown in~\fig{fig:xela_architecture}.

The student and teacher share the same encoder and projector head architecture, both initialized with the same weights. The projector head corresponds to a 3-layer MLP with an output dimension of $k=65536$. We use the projection head for the proxy prediction task to distill knowledge to match output distributions over $k$ dimensions between student and teacher networks. The student network is updated via back-propagation, while the teacher network is updated at a lower frequency via exponential moving average (EMA) on the student weights. We pass the global and local views to the student encoder, while the teacher only has access to the global views. The register tokens from global/local views are passed through the projection head. For the teacher only, the output is also centered and sharpened via softmax normalization.

\section{Additional task details}

We provide additional information about the decoder architectures for each task, as well as additional results to highlight the performance on downstream tasks when using frozen or fine-tuned \model representations. Also, please refer to Table~\ref{tab:datasets_bench} for details on labeled data curation for evaluation tasks.

\begin{table*}[htbp]
\centering
\resizebox{\textwidth}{!}{
\begin{tabular}{lcccc}
\hline
\multicolumn{1}{c}{Task} & Dataset                          & Size                 & Collector            & Label               \\ \hline
\multirow{2}{*}{Force estimation} &
  \multirow{2}{*}{\begin{tabular}[c]{@{}c@{}}Normal load\\(indenter: sphere, flat)\end{tabular}} &
  50k datapoints &
  Robot &
  3-axis force  \\ \\
  
Pose estimation & Object sliding                     & 108 trajectories                  & Human                & \begin{tabular}[c]{@{}c@{}}Object pose $\mathbf{SE}(2)$\end{tabular}           \\ [0.7mm]

Joystick state estimation & Joystick motion                     & 817 trajectories                  & Human                & \begin{tabular}[c]{@{}c@{}}Normalized roll, pitch, yaw \end{tabular}           \\ [0.7mm]

Plug insertion & Demonstrations & 100 trajectories & Human & Absolute EE pose\\[0.7mm]

\hline
\end{tabular}
}
\caption{Datasets for evaluating \model representations on downstream tasks.}
\label{tab:datasets_bench}
\end{table*}

\subsection{Force estimation}
\model features are pooled via attentive pooling to obtain a full-hand representation $z_{hand} \in \mathbb{R}^d $. The force decoder consist of shallow 2-layer MLP with 3 outputs regressing to normalized force for each axis.

\begin{figure}[ht]
    \centering
    \begin{subfigure}[ht]{0.49\linewidth}
        \centering
        \includegraphics[width=\textwidth]{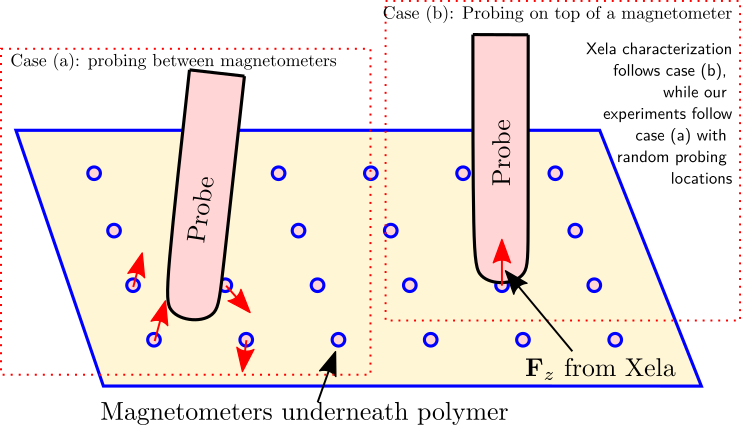}
        \caption{Illustration of data collection procedure followed in our setup vs procedure followed during Xela force calibration}
    \end{subfigure}
    \begin{subfigure}[ht]{0.49\linewidth}
        \centering
        \includegraphics[width=\textwidth]{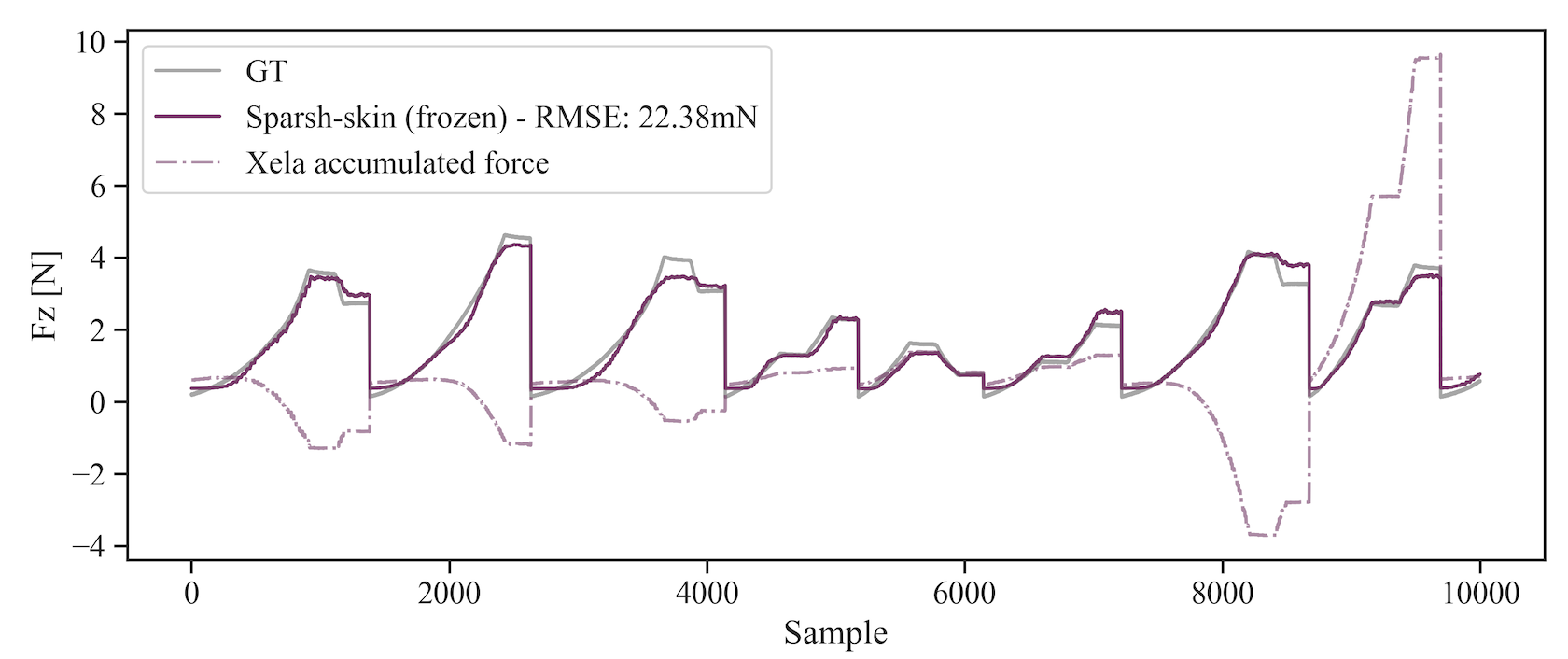}
        \caption{\small{Illustration of how the Normal Xela force output is inconsistent with GT force obtained from F/T probe, and the corresponding prediction from \model}}
    \end{subfigure}
    \caption{Illustration of data collection protocol follwed for Force estimation with Xela sensors}
    \label{fig:xela_force_clarification}
    \vspace{-0.5cm}
\end{figure}

In Figure~\ref{fig:xela_force_clarification} we illustrate the data protocol followed for force estimation, which we note is different from the protocol that is usually followed for force characterization of tactile sensors. We note that we indent the tactile sensor pads at both, positions on top of the sensor as well as positions in between magnetometer locations, while choosing these positions randomly. This results in cases where the probe may slide and present slightly uneven force outputs. Specifically, in figure~\ref{fig:xela_force_clarification}(b) we note that \model predicts the correct normal forces, while accumulation (mean) of normal forces from the magnetometers over the sensor pad results in inconsistent force outputs compared to ground truth.

\begin{figure}[htbp]
    \centering
    \includegraphics[width=\linewidth]{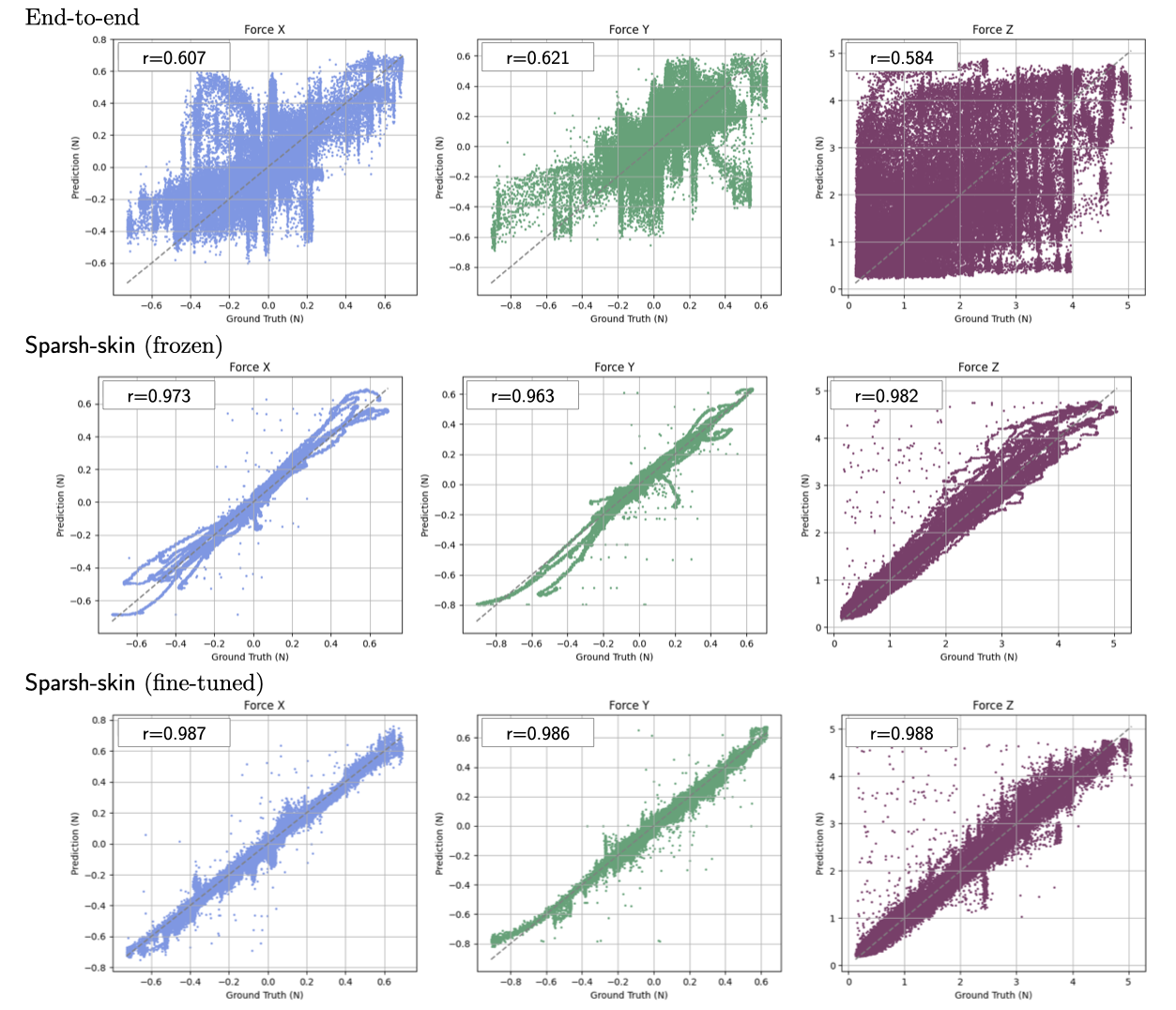}
    \caption{Correlation between ground truth and predicted forces on unseen normal loading with an indenter on Xela sensors.}
    \label{fig:force_correlation}
\end{figure}

In \fig{fig:force_correlation}, we present the correlation metrics between ground truth and predicted forces on test data for decoders trained with a 33\% data budget. The results show that end-to-end training leads to overfitting, resulting in poor generalization to unseen strokes and essentially random normal force predictions. In contrast, using \model (frozen) representations yields better fitting, which can be further improved by adapting these representations to in-domain data.

\begin{figure}[htbp]
    \centering
    \includegraphics[width=0.8\linewidth]{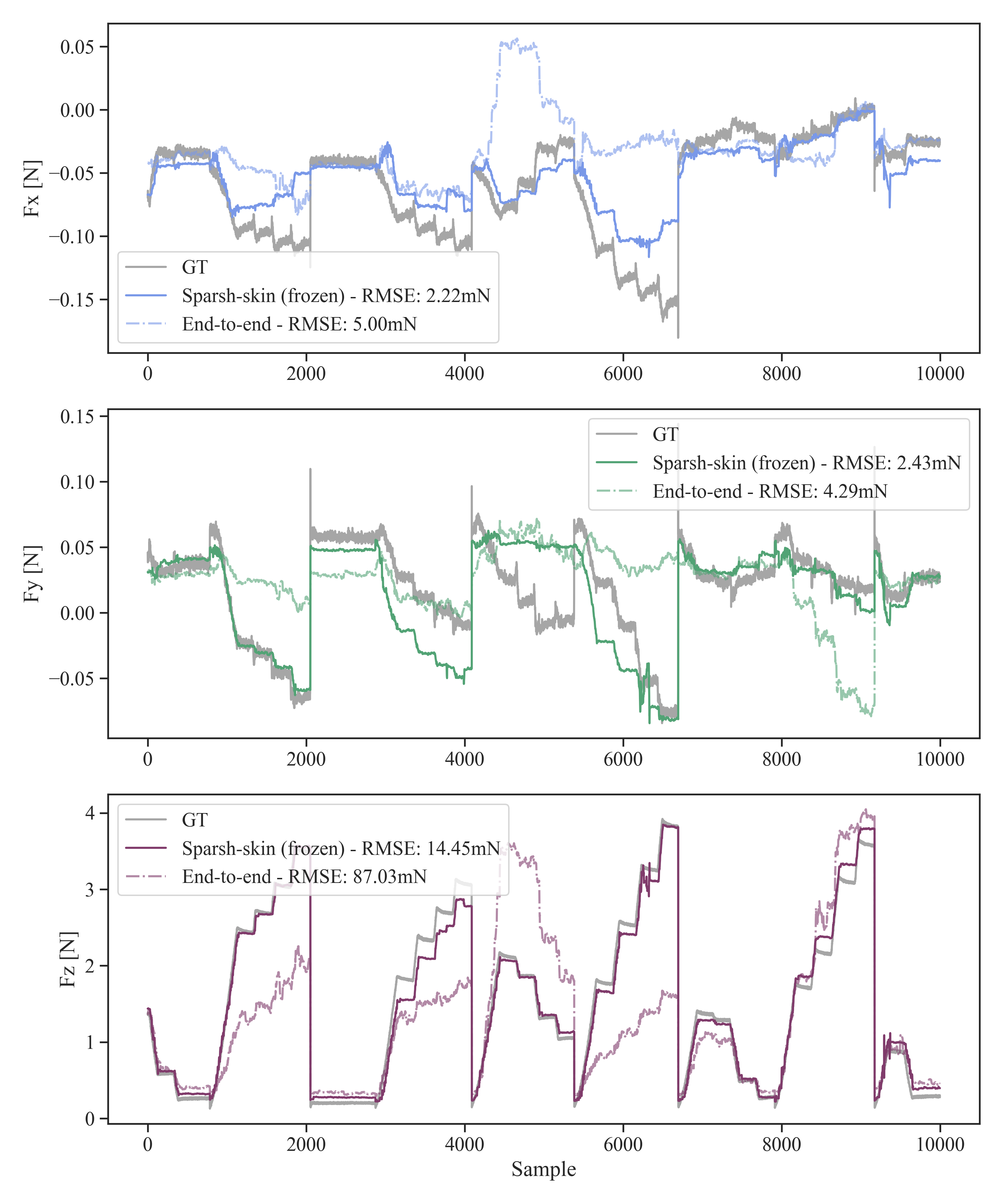}
    \caption{Ground truth tangential and normal force from test strokes with flat indenter (gray) and force sequence reconstruction from \model (frozen) and end-to-end model.}
    \label{fig:force_strokes}
\end{figure}

In \fig{fig:force_strokes}, we present a comparison between ground-truth testing strokes (normal loading sequences) and their reconstructed counterparts, obtained by passing Xela data through the frozen force decoder to recreate the sequences. The forces estimated via \model (frozen) are able to capture increasing/ decreasing changes in the normal loading, as opposed to the end-to-end model. Shear from skin representations is not as accurate as normal force prediction, but the trend of the tangential forces matches the ground truth.

\subsection{Joystick state estimation} 
\begin{figure}[htbp]
    \centering
    \includegraphics[width=0.75\linewidth]{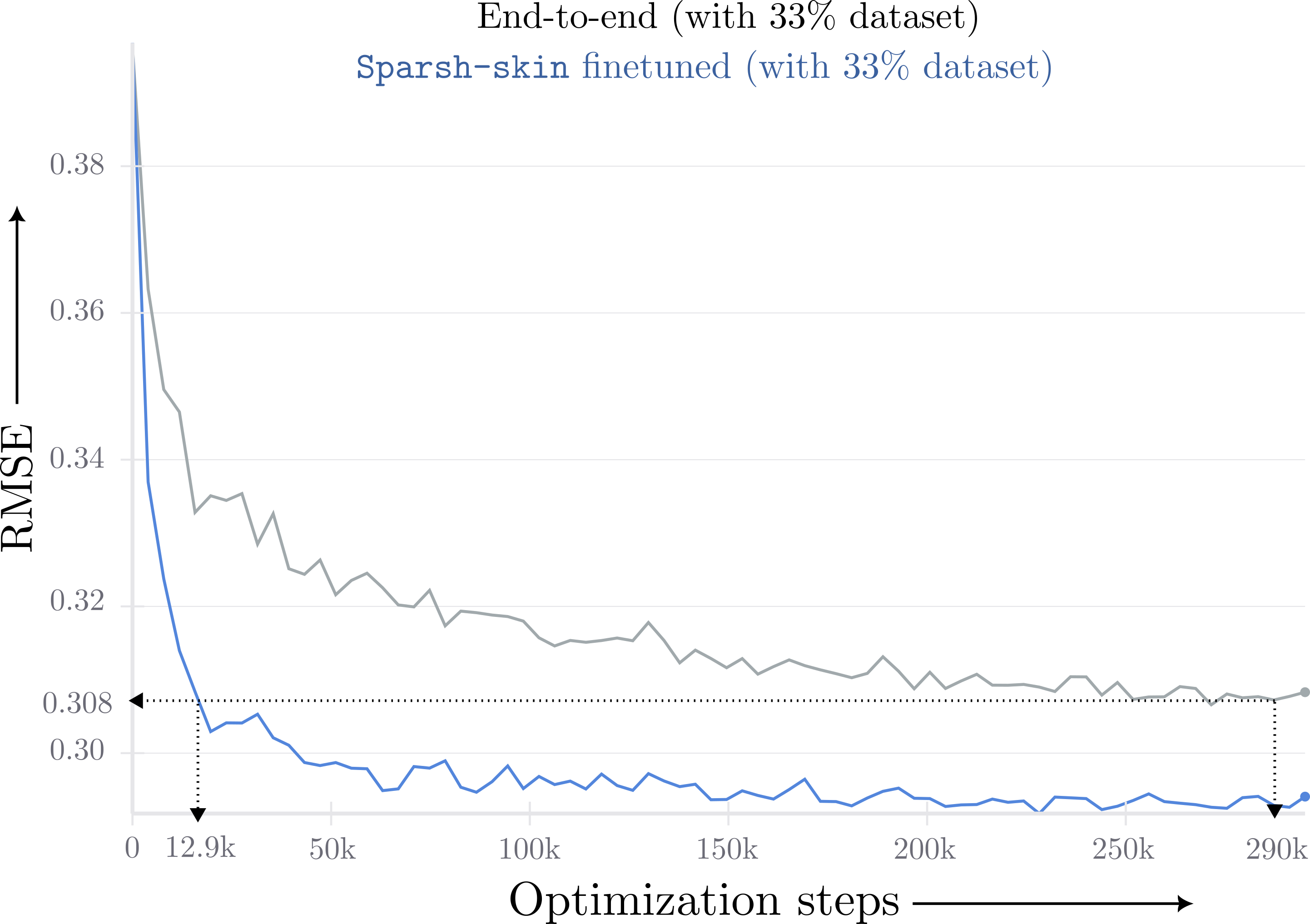}
    \caption{Validation RMSE convergence rates between \model fine-tuned and \model end-to-end: We find that \model fine-tuned allows the model to generalize and learn the patterns required to infer joystick states significantly faster during training.}
    \label{fig:joystick_convergence}
\end{figure}

For this task, we highlight that when we train decoders using pretrained representations as the input, the convergence rate of the validation RMSE is significantly higher (see \fig{fig:joystick_convergence}) than training the decoder using raw observations through uninitialized models. Specifically observe that \model (fine-tuned) is able to reach performance on par with end-to-end pretrained model within 12.9k optimization steps.

\subsection{Pose estimation}
In this task, we aim to predict the object pose over 1-second trajectories. Xela observations at 100Hz are converted into tactile representation tokens at the output frequency using \model in a cascaded manner. Following attentive pooling, a single-layer transformer block is applied to reason about the 1-second context window of full-hand tactile features.

\begin{figure*}[htbp]
    \centering
    \includegraphics[width=\linewidth]{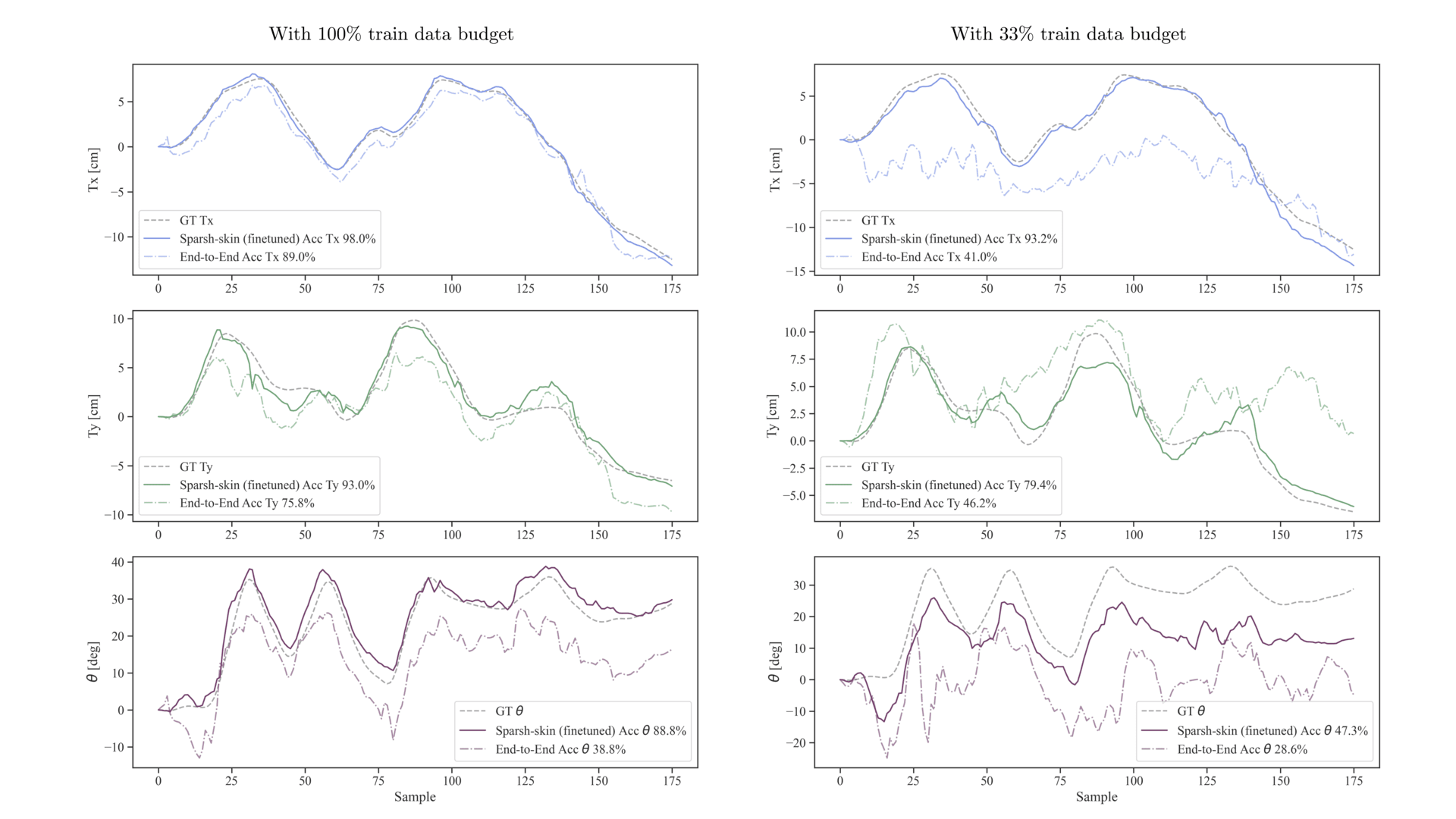}
    \caption{Ground truth pose sequence for object in test set and reconstructed trajectory via end-to-end and \model (finetuned) representations. (left) Task decoders trained with 100\% of train data budget, corresponding to 108 sequences. (right) Task decoders trained with 33\% of train sequences.}
    \label{fig:pose_sequences}
\end{figure*}

\begin{figure}[ht]
    \centering
    \includegraphics[width=\linewidth]{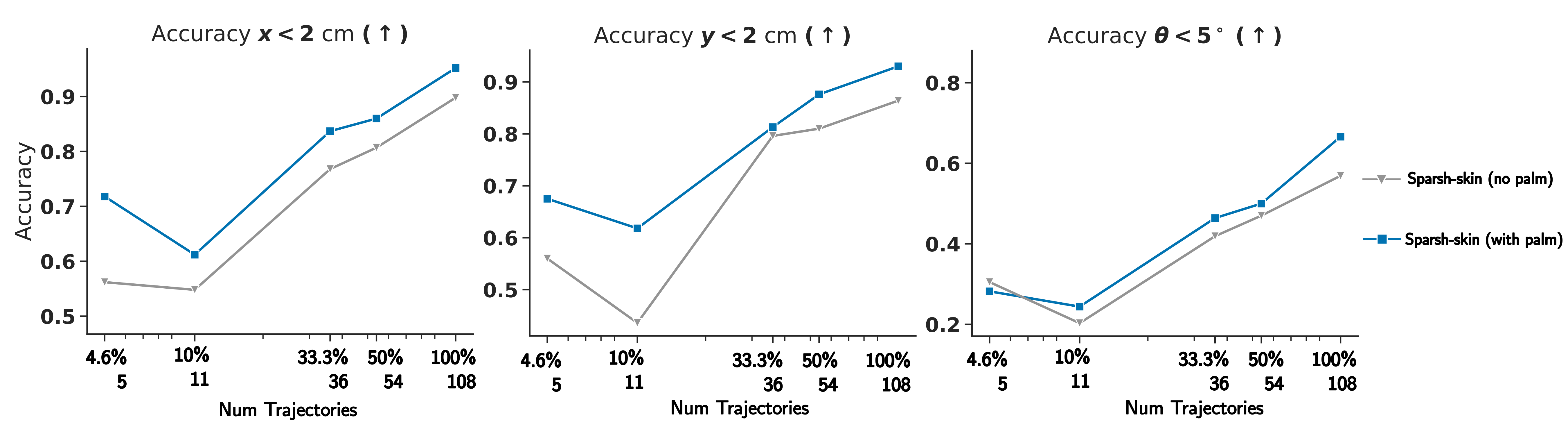}
    \caption{\small{Comparison of pose estimation accuracy of \model with and without palm sensing.}}
    \label{fig:pose_estimation_palm_nopalm}
    \vspace{-0.4cm}
\end{figure}

\fig{fig:pose_sequences} compares ground-truth test pose sequences with their reconstructed counterparts, obtained from task models trained on 100\% and 33\% of the available data. The results show that fine-tuning \model on the full dataset yields higher accuracy in estimating object pose changes over time compared to traditional end-to-end approaches. Moreover, even with a drastic reduction in labeled samples (to 33\%), the model still achieves relatively good performance, particularly in tracking translation changes. Furthermore, for this task, we also visualize that this tasks requires \emph{full-hand} sensing. For instance, in \fig{fig:pose_estimation_palm_nopalm}, we observe that when we use \model by removing palm sensing on the Xela hand results in $\geq 10\%$ drop in pose tracking performance.

\subsection{Policy learning (plug insertion)} 
\begin{figure}[htbp]
    \centering
    \includegraphics[width=\linewidth]{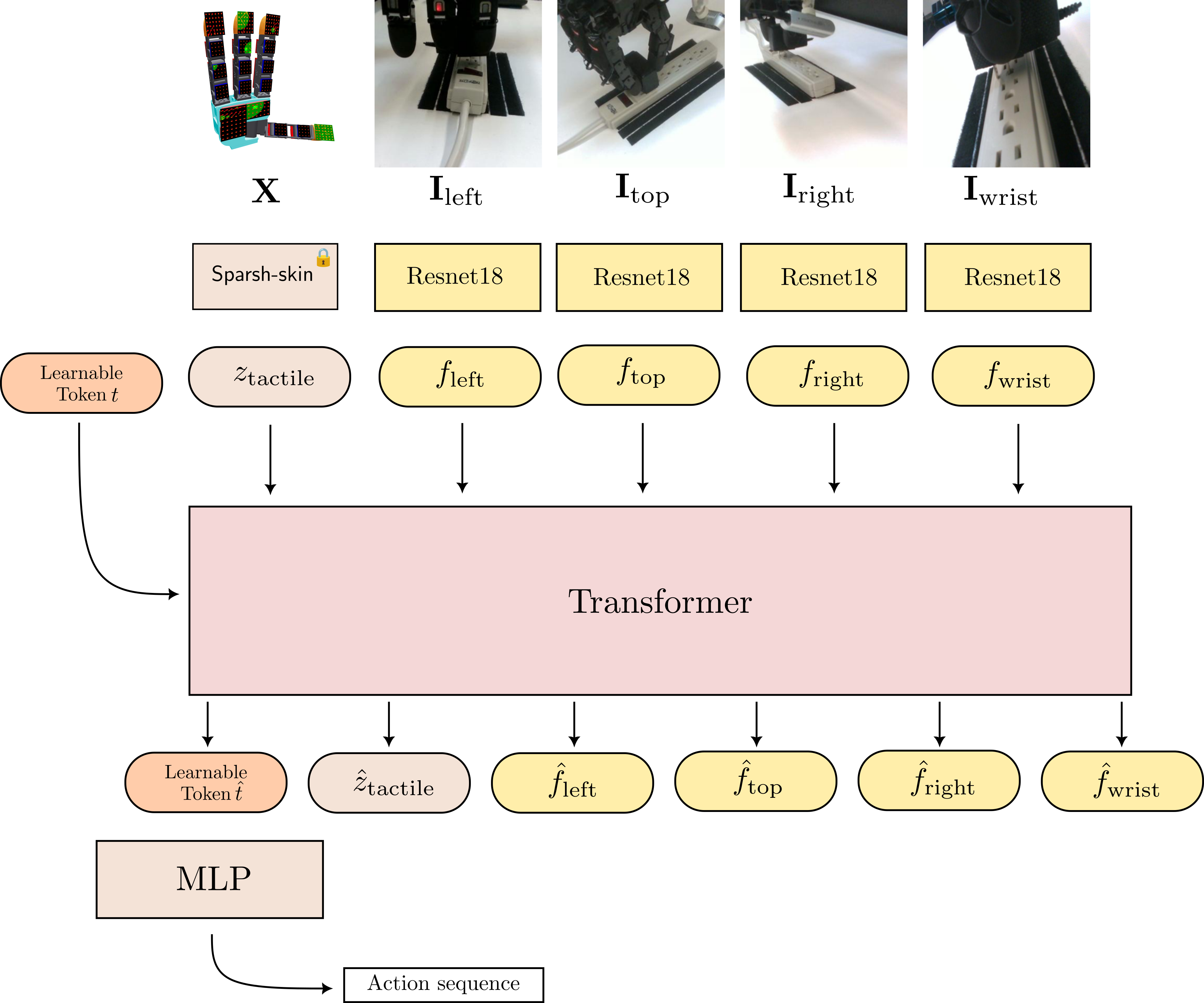}
    \caption{Illustration of the policy architecture: We use a transformer to fuse information from visual and tactile modalities, through the use of a learnable action token, which is then used to subsequently predict action sequences.}
    \label{fig:policy_decoder}
\end{figure}

For this task, we use a transformer decoder to predict action sequences given camera and tactile observations. \fig{fig:policy_decoder} illustrates the architecture of the transformer decoder used in this work. Images are encoded using a Resnet18 CNN, which are trained from scratch to produce image features, while the tactile observations are processed through \model. Further, a learnable token (CLS / action) token is also concatenated with the observation tokens.
After processing through the transformer, we extract the action token, which is then passed into a small 2-layer MLP to predict a  sequence of actions. For this task, follow an \emph{receding-horizon} control approach, where we choose a prediction action sequence length of 16, of which 8 actions are executed, given only the observations from the current timestep.

\end{document}